\documentclass[sn-mathphys,Numbered]{sn-jnl}% Math and Physical Sciences Reference Style
\usepackage{graphicx}%
\usepackage{multirow}%
\usepackage{amsmath,amssymb,amsfonts}%
\usepackage{amsthm}%
\usepackage{mathrsfs}%
\usepackage[title]{appendix}%
\usepackage{xcolor}%
\usepackage{textcomp}%
\usepackage{manyfoot}%
\usepackage{booktabs}%
\usepackage{algorithm}%
\usepackage{algorithmicx}%
\usepackage{algpseudocode}%
\usepackage{listings}%
\usepackage{amsmath,amsfonts}
\usepackage{array}
\usepackage{stfloats}

\usepackage{subcaption}
\usepackage{graphicx}
\usepackage{siunitx}
\usepackage[english]{babel}
\usepackage{fancyhdr}
\usepackage{transparent}
\usepackage{makecell}
\usepackage{multirow}

\theoremstyle{thmstyleone}%
\theoremstyle{thmstyletwo}%

\theoremstyle{thmstylethree}%

\begin{document}

\title[Article Title]{Separability and Scatteredness (S\&S) Ratio-Based Efficient SVM Regularization Parameter, Kernel, and Kernel Parameter Selection}


\author*[1]{\fnm{Mahdi} \sur{Shamsi}}\email{mahdi.shamsi@torontomu.ca}

\author[1]{\fnm{Soosan} \sur{Beheshti}}\email{soosan@torontomu.ca }
\equalcont{These authors contributed equally to this work.}


\affil*[1]{\orgdiv{Department of Electrical Computer and Biomedical Engineering}, \orgname{Toronto Metropolitan University}, \orgaddress{\street{350 Victoria St}, \city{Toronto}, \postcode{M5B 2K3}, \state{Ontario}, \country{Canada}}}

% \affil[2]{\orgdiv{Department}, \orgname{Organization}, \orgaddress{\street{Street}, \city{City}, \postcode{10587}, \state{State}, \country{Country}}}

% \affil[3]{\orgdiv{Department}, \orgname{Organization}, \orgaddress{\street{Street}, \city{City}, \postcode{610101}, \state{State}, \country{Country}}}


\abstract{Support Vector Machine (SVM) is a robust machine learning algorithm with broad applications in classification, regression, and outlier detection. SVM requires tuning the regularization parameter (RP) which controls the model capacity and the generalization performance. Conventionally, the optimum RP is found by comparison of a range of values through the Cross-Validation (CV) procedure. In addition, for non-linearly separable data, the SVM uses kernels where a set of kernels, each with a set of parameters, denoted as a grid of kernels, are considered. The optimal choice of RP and the grid of kernels is through the grid-search of CV. By stochastically analyzing the behavior of the regularization parameter, this work shows that the SVM performance can be modeled as a function of separability and scatteredness (S\&S) of the data. Separability is a measure of the distance between classes, and scatteredness is the ratio of the spread of data points. In particular, for the hinge loss cost function, an S\&S ratio-based table provides the optimum RP. The S\&S ratio is a powerful value that can automatically detect linear or non-linear separability before using the SVM algorithm. The provided S\&S ratio-based table can also provide the optimum kernel and its parameters before using the SVM algorithm. Consequently, the computational complexity of the CV grid-search is reduced to only one time use of the SVM. The simulation results on the real dataset confirm the superiority and efficiency of the proposed approach in the sense of computational complexity over the grid-search CV method.}


\maketitle

\section{Introduction}
Support Vector Machine (SVM) is a well-known supervised learning algorithm for linear and nonlinear data classification, regression \cite{gao2002probabilistic}, and outlier detection \cite{lenz2022optimised, widodo2007support,byvatov2003support} in a wide range of applications, such as face detection \cite{dino2019facial, kumar2019face}, bio-informatics \cite{daberdaku2019antibody, sanz2018svm} and text categorization \cite{dhar2021text, berge2019using}. The SVM kernel trick method enables the algorithm to perform nonlinear classification by projecting the data points into a higher-dimensional feature space and using the linear classification for this feature space.

The strength of SVM is in its independence of the generalizability of the model from the dimension of the input space due to a strong theoretical background in learnability and the Vapnik-Chervonenkis (VC) dimension \cite{blumer1989learnability}. However, the generalization depends on the choice of a hyperparameter, known as the regularization parameter\cite{suykens1999least, wang2005support} that determines the weights of misclassified data points in the cost function. The poor choice of this hyperparameter can drastically decrease generalization performance, while its optimal choice helps the SVM model avoid overfitting and underfitting problems \cite{hastie2004entire}. Usually, the value of the regularization parameter is set to a recommended value by an expert in the field. The conventional method for choosing the regularization parameter of the SVM model is the k-fold cross-validation (CV) method \cite{cherkassky2007learning}. There are two major drawbacks to the Cross-Validation (CV) method. First, model training must be rerun $k$ times from scratch, and therefore this method is computationally expensive and time consuming; second, the final results are sensitive to the initial range of parameter values \cite{hastie2004entire}.  Consequently, while grid search cross-validation is a widely used strategy for hyperparameter optimization in machine learning, the cost of the algorithm increases exponentially with the number of hyperparameters, especially in multi-class SVM with kernel trick\cite{bergstra2012random, hsu2003practical, noble2006support}.

 In this paper, we propose a stochastic-based approach for selecting the optimal regularization parameter, leveraging the hinge loss and the underlying statistical properties of the data to estimate the optimal value of the regularization parameter. The proposed method reduces the computational complexity associated with hyperparameter selection in the SVM algorithm and shows additional advantages in predicting the performance of SVM. The method uses a statistical analysis of the performance of Hinge loss. 
 Hinge loss plays a vital role in SVMs as it quantifies the misclassification error with respect to the margin width. It is used as the loss function in the optimization problem that SVMs aim to solve. 
By minimizing the hinge loss, SVMs attempt to find the best separating hyperplane between classes, taking into account the trade-off between margin maximization and classification error. 

 The proposed method focuses on the relationship between hinge loss, the regularization parameter, and the statistical properties of the data, and introduces new notations, specifically separability and scatteredness (S\&S) for the analysis. Separability is the Euclidean distance between the true centers of the two classes and serves as a measure of how distinct the classes are in the input space. Scatteredness is the ratio of the spread of data points in the two classes, defined as the average standard deviation of all data points, providing an indication of the degree of overlap between classes. The proposed approach is grounded in a deep understanding of the statistical behavior of SMV based on the regularization parameter, the SVM's margin width, hinge loss, separability, and scatteredness of the input data. Consequently, the structural model for the data which incorporates the underlying probability distribution of the data of the classes provides an optimal SVM model. The statistical approach is also able to choose the optimal kernel and its associated parameters. Kernel functions play a critical role in the performance of SVMs by allowing the transformation of data points into a higher-dimensional space where they can be more easily separated. The choice of the appropriate kernel function, as well as its parameters, is essential for the success of the SVM model. In this work, we propose a novel analytical method to determine the optimal kernel function and the corresponding parameters by calculating the S\&S in the feature space resulting from the kernel transformation. By doing so, we take into account the intrinsic structure and geometry of the transformed data points, allowing for a more informed and effective selection of the kernel and its parameters. The proposed approach shows the important role of the S\&S in classification that is analogous to the role of the signal-to-noise ratio in the analysis of errors in data modeling. One strength of the approach is in automating the test for linear separability. It is shown that when the calculated S\&S is below -5dB, then the data is not linearly separable and the kernel method should be used. 
 
The paper is organized as follows: Section \ref{sec:MathBack} provides a brief mathematical background of the SVM algorithm and discusses the importance of the regularization parameter in the generalizability of SVM models. Section \ref{sec:Proposed} introduces two important definitions that are used for the stochastic analysis of the regularization parameter in Section \ref{sec:sandspr}. Section \ref{sec:Kernel} extends the proposed S\&S based method for the kernel method and multiclass classification in the SVM algorithm. Finally, the simulation results are presented in Section \ref{section:simu}. 

\section{SVM Binary Classification}
\label{sec:MathBack}
\begin{figure}[hbt!]
\begin{center}
\includegraphics[clip,trim=0cm 0cm 0cm 0cm,scale=0.4]{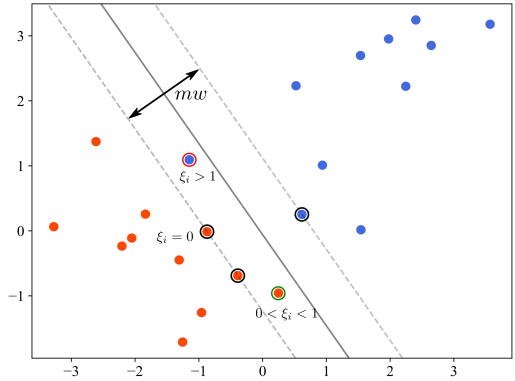}
\caption{Soft-margin SVM}\label{fig:fig3}
\end{center}
\end{figure}

In binary classification, machine learning algorithms receive a data set $S=((\mathbf{x}_1,y_1), \dots, (\mathbf{x}_n,y_n))$ as input and produce an output in the form of a hypothesis or prediction rule. The data points or feature vectors denoted as $\mathbf{x}_i, i \in \{1, \dots, n \}$ are $\psi$-dimensional vectors ($\mathbf{x}_i \in \mathbb{R}^\psi$), which are assumed to be samples of a domain set, and the target values $y_i \in \{-1,+1 \}$, $i \in \{1,\dots,n \}$, are the respective labels. Positive data points associated with $y_i=+1$ are indicated by $\mathbf{x}_+$, and negative data points associated with $y_i=-1$ are indicated by $\mathbf{x}_-$. The prediction rule then receives unlabeled data points from the domain set and outputs the labeled data point. In the binary classification task, the prediction rule results in a decision boundary between positive and negative data points (separating the hyperplane). The goal of the hard-margin support vector machine algorithm is to maximize the margin that separates the data points of these two classes:
\begin{gather}
    \underset{\mathbf{w},b}{\min} f(\mathbf{w}) = \frac{1}{2}||\mathbf{w}||^2 \\
    \mathrm{s.t.}\;\;g(\mathbf{w},b)=y_i(\mathbf{x}_i^T\mathbf{w}+b) \geq 1 \label{eqn:HRsvmO}
\end{gather}
The issues of nonseparable datasets and outlier sensitivity are considered in the support vector machine (SVM) framework by incorporating slack variables \cite{sain1996nature, scholkopf2002learning}. If the data points in the two classes overlap, as shown in Figure \ref{fig:fig3}, there is no hyperplane that separates the two classes. In this scenario, the SVM classification conditions can be relaxed by allowing some data points to violate the margin constraints while penalizing them \cite{Feynman1963118}. As a result, the objective function of the SVM algorithm is:
\begin{align}
\label{eqn:marginalsvm}
    \underset{\mathbf{w},b, \xi}{\min} &f(\mathbf{w}) = \frac{1}{2}||\mathbf{w}||^2 + C\sum_{i=1}^{n} \xi_i\\
    \mathrm{s.t.}\;\;&y_i(\mathbf{x}_i^T\mathbf{w}+b) \geq 1-\xi_i, \;\; (i=1,\dots,n) \nonumber \\
    & \xi_i \ge 0, \quad (i = 1,\ldots,n)
\end{align}
\noindent where $\xi_i$'s are slack variables associated with their corresponding $x_i$ and the regularization parameter $C > 0$ (RP-$C$) is a user-defined regularization parameter that controls the trade-off between maximizing the margin and minimizing the classification error, which controls the width of the margin (shown in Figure 1 as $mw$).
When comparing the SVM cost function in (\ref{eqn:HRsvmO}) with that of the soft SVM in (\ref{eqn:marginalsvm}), an additional term of the slack function in the form of $C\sum_{i=1}^{n} \xi_i$ is also minimized. However, compared to the SVM constraint in (\ref{eqn:HRsvmO}), the constraint of soft SVM in (\ref{eqn:marginalsvm}) allows the margin to be less than 1 while penalizing any data point that falls within the margin with $C\xi_i$. Note that when the regularization parameter $C$ approaches infinity, the SVM model becomes very sensitive to constraint violations (misclassification). The regularization parameter $C$ controls the trade-off between maximizing the margin (minimizing $\lVert \mathbf{w} \rVert^2$) and minimizing the sum of the slack variables $\sum_{i=1}^n \xi_i$. A large value of $C$ places greater emphasis on minimizing classification errors and constraint violations, resulting in a smaller margin. This can lead to a model with higher complexity that fits the training data very closely. However, the model may not generalize well to unseen data as it becomes sensitive to noise and outliers, resulting in overfitting. A small value of $C$ places greater emphasis on maximizing the margin and less on minimizing classification errors, allowing for some constraints violations. This results in a model with lower complexity and wider margin. Although the model may not fit the training data as closely, it is more likely to generalize well to unseen data, avoiding overfitting.

\subsection{Choice of Optimal RP-$C$}
The regularization parameter, $C$, is a hyperparameter that is not learned during the training process and has to be provided directly to the algorithm \cite{hastie2004entire}. The choice of $C$ significantly impacts the generalization performance of the SVM model. Therefore, it is essential to choose an appropriate value for $C$ to achieve a balance between model complexity and generalization performance.
Conventionally, to find the optimum value of $C$, a cost function or a scoring metric is considered to evaluate the performance of the model. For each choice of $C$, solving the SVM optimization problem leads to the estimated $\mathbf{w}$ and $b$: 
\begin{equation}
    C \rightarrow (\mathbf{w}_c,b_c)
\end{equation}
Consequently, the desired cost function denoted by $L(C,S)$ can be calculated as a function of these estimated parameters and the training data set $S$. The optimal regularization parameter denoted by $C_{opt}$ is:
\begin{align}
    C_{opt} = \underset{C \in \left [C_{\min}: \Delta_C: C_{\max}\right ]}{\arg \min}L(C,S) \label{eqn:coptcost}
\end{align}
where $\left [C_{\min}: \Delta_C: C_{\max}\right ]$ denotes a grid of $C$ values between $C_{\min}$ and $C_{\max}$ with step size $\Delta_C$.
A common method that utilizes (\ref{eqn:coptcost}) to determine $C$ is cross-validation \cite{allen1974relationship, stone1974cross, duarte2017empirical}.  Grid search cross-validation \cite{cherkassky2007learning} method selects the optimal tuning parameters from a one-dimensional or multidimensional grid during this search, where all possible combinations of hyperparameters are used to train the model \cite{cherkassky2007learning, chang2011libsvm}. 

\section{Optimal Regularization Parameter as Function of Separability and Scatteredness}
\label{sec:Proposed}
\begin{figure}[hbt!]
\begin{center}
\includegraphics[scale=0.3]{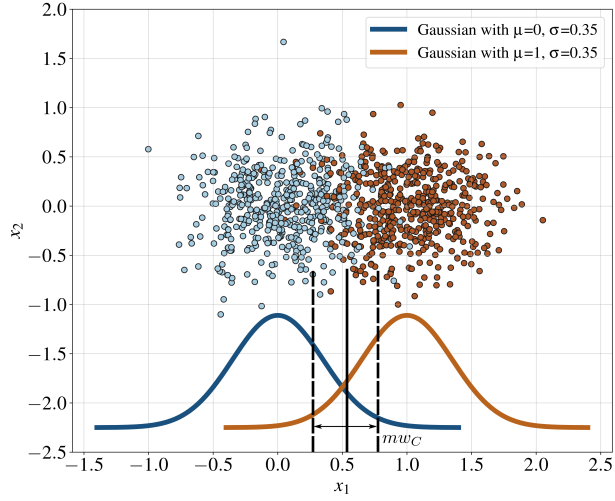}
\caption{Two classes are generated from Gaussian distributions with $\mu_1 = (0, 0)$ and $\mu_2 = (0, 1)$ and $\sigma_1=0.35=\sigma_2=0.35$.}\label{fig:pdf}
\end{center}
\end{figure}

In this section, to analyze the behavior of the optimal choice of RP-$C$, two data clusters with two-dimensional features are generated from two Gaussian distribution. The centers (means) of the two classes are $\mu_1$ and $\mu_2$ each with standard deviations $\sigma_1$ and $\sigma_2$ respectively. Figure \ref{fig:pdf} shows 1000 sample data points for each class with $\mu_1=(0,0)$ and $\mu2=(1,0)$ and standard deviations $\sigma_1=0.35=\sigma_2=0.35$.  

Before getting into the details of the proposed method, three important definitions are introduced in the following section.
\subsection{Important Definitions}

\textbf{Separability:} is the Euclidean distance between the true center of the two classes, which is denoted by $d$ which is the distance between $\mu_1$ and $\mu_2$:
\begin{align}
    d = |\mu_1 - \mu_2|
\end{align}

\textbf{Scatteredness:} is the average of the spread of data points in the two classes which is the average standard deviation of all data points and is denoted by $\sigma$
\begin{eqnarray}\label{eq:29}
\sigma = \sqrt{\frac{{n_1} {\sigma^2_1}+{n_2} {\sigma^2_2}}{n_1+n_2}}
\end{eqnarray}
The directional scatterdness can be calculated by first projecting the data points onto the line between the centers. This value is helpful in analyzing the degree of separation and overlap between the two classes.

\textbf{Separability and Scatterdness (S\&S) Ratio:} Inspired by the notion of the Signal-to-Noise ratio (SNR) in signal processing, the concept of Separability and Scatteredness (S\&S) ratio is defined as
\begin{eqnarray}\label{eq:30}
& \mathrm{S\&S}\;\; Ratio=20\log _{10} \frac{d}{\alpha \times \sigma}
\end{eqnarray}
where $\alpha$ is a normalizing factor and can be chosen by the user. In this work, we chose $\alpha$ to be $6=3\time3$ according to the 3-$\sigma$ rule. For the normally distributed data points, we can expect that 99.7\% of the data will be within three standard deviations of the mean, with only 0.3\% of the data falling outside this range. This choice is a good indicator of the relative separability of the two data sets. 

\subsection{Optimal RP-$C$ as a Function of Seperability and Scatterdness}
\begin{figure}[hbt!]
\begin{center}
\includegraphics[scale=0.32]{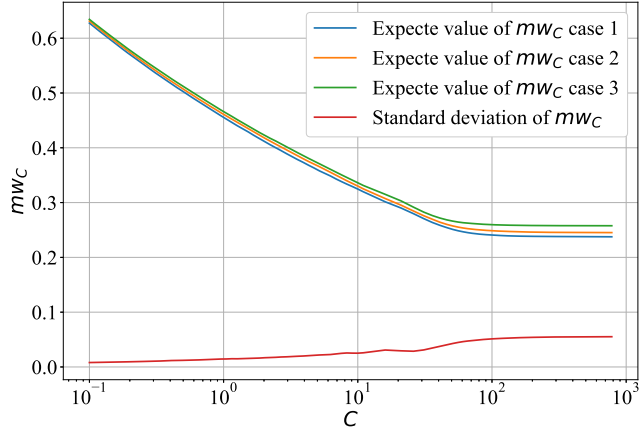}
\caption{Expected value and standard deviation of margin width of SVM with respect to the range of RP-$C$ values (for $\sigma=0.12$)}\label{fig:margin}
\end{center}
\end{figure}

In the considered binary setting that is generated by the two Gaussian distribution, any choice of $C$ generates a unique margin width denoted by $mw_C$ which is deterministic:
\begin{align}
    mw_C=f(\mu_1,\mu_2,\sigma_1,\sigma_2)
\end{align}
Figure \ref{fig:margin} shows the behavior of the mean and standard deviation of the generated margin width with respect to RP-$C$ for three scenarios. These values are estimated by averaging over 1000 runs. Three different cases with the same scatteredness factor $\sigma=0.12$,  defined in (\ref{eq:29}), are generated as follows:

Case 1: The size of each class is $n1 = n_2 = 2000$, with standard deviations $\sigma_1 = \sigma_2 = 0.12$ for both classes.

Case 2: The sizes of the first and second classes are $n_1 = 1000$ and $n_2 = 2000$, respectively, with standard deviations $\sigma_1 = 0.09$ and $\sigma_2 = 0.132484$.

Case 3: The size of the first and second classes are $n_1 = 1000$ and $n_2 = 2500$ respectively, with standard deviations $\sigma_1 = 0.16$ and $\sigma_2 = 0.099600$.

As the figure shows, the margin width is mainly a function of the scatterdness $\sigma$. 
On the other hand, if the structure of the validation cost function in (\ref{eqn:coptcost})  is such that it is an average of pointwise values, i.e.,
\begin{align}
   L(C,S)= \frac{1}{n}\sum L(C,x_i,y_i)  \label{eqn:ValCost}
\end{align}
Then the expected value and variance of this cost function as a function of $C$ are
\begin{align}
    &E(L(C,S))= E( L(C,x_i,y_i)) \\
    &var(L(C,S))= \frac{1}{n^2} var ( L(C,x_i,y_i))
\end{align}
As the variance of this value is expected to be much smaller than that of its mean, the samples of this cost function generated by the available data perform almost identical to the expected value. Also, due to the central limit theorem, the cost function distribution for any $C$ behaves like a Gaussian distribution regardless of the distribution of the classes. Note that while for simplicity we use the Gaussian distribution for the two classes in this analysis, the argument can be generalized for any class distribution.

% Figure \ref{fig:Hingloss_all} shows the hing loss for test and train data set  in (\ref{eqn:Coptwithsigma12}) with respect to the hinge loss for $\sigma_1 = \sigma_2 = \sigma$ by averaging over 1000 runs. 

The behavior of the expected value of the cost function for two Gaussian distribution classes is only a function of the $C$ value and the variance and mean of the two distributions. Assuming that the mean of one class is at zero ($\mu_1=0$), this value will be a function of the standard deviations and the distance between the two means $d=|\mu_1-\mu2|$: 
\begin{align}
    E(L(C,S, \sigma_1, \sigma_2, d)) = \frac{n_1}{n} E(L(C, S_+, \sigma_1, \sigma_2, d)) + \frac{n_2}{n} E(L(S_-, \sigma_1, \sigma_2, d))
\end{align}
where $S_+$ is the set of data for which $y$ value is +1 and has $n_1$ elements and $S_-$ is the set of data for which $y$ value is -1 and has $n_2$ elements. Normalizing the distance between the two Gaussian distributions will change the margin width to $mw_C/d$ and, accordingly, changes the cost function as follows: 
 \begin{align}
     E(L(C, S_+, \sigma_1, \sigma_2, d)) &= \nu E(L(C, S_+, \frac{\sigma_1}{d}, \frac{\sigma_2}{d}, 1)) \label{eqn:Ekappa1}\\
     E(L(C, S_-, \sigma_1, \sigma_2, d)) &= \nu E(L(C, S_-, \frac{\sigma_1}{d}, \frac{\sigma_2}{d}, 1)) \label{eqn:Ekappa2}
 \end{align}
where $\nu$ is a scaling value that is not a function of $C$. Therefore, the optimum $C$ in (\ref{eqn:coptcost}) can be alternatively calculated as follows.
\begin{align}
    C_{opt}(\sigma_1, \sigma_2, d)= \underset{C}{\arg \min}\; E(L(C, S, \frac{\sigma_1}{d}, \frac{\sigma_2}{d}, 1)) \label{eqn:Coptwithsigma12}
\end{align}
Therefore, this value of $C_{opt}$ can be calculated for a range of values of $\sigma_1$, $\sigma_2$ and $d$ and can be used as a lookup table. This can replace the need for the grid search in (\ref{eqn:coptcost}) for any binary classification.  For any given binary scenario, it is only needed to find the associated values $\sigma_1$ and $\sigma_2$ and $d$ and then read the value of the optimum $C$ from the available table.
\section{S\&S Ratio-Based Hinge Loss Optimum RP-$C$} \label{sec:sandspr}
Hinge loss measures the degree to which a data point violates the margin of the SVM classifier, which is a key component of the optimization problem solved by SVM to find the optimal decision boundary (hyperplane) that separates classes in the feature space. The hinge loss per data point is calculated as:
\begin{align}
\label{eqn:hingloss}
    L_H(C,\mathbf{w}, b, \mathbf{x}_i, y_i)=\max(0, 1 - y_i(\mathbf{w}^T\phi(\mathbf{x}_i) + b))
\end{align}
Therefore, the hinge loss, that is, the summation of these per-point elements, is a lower bound for the summation of slack variables $\xi_i$s:
\begin{equation}
\sum_{i=1}^{n}\xi_i \ge L_H(C,\mathbf{w}, b, \mathbf{x}_i, y_i)
\end{equation}
The hinge loss associates penalties for misclassified data points based on their distance from the decision boundary. Therefore, data points that are farther away from the decision boundary have a greater loss.
Hinge loss is less sensitive to outliers compared to other loss functions, such as squared loss. In SVM, outliers do not significantly affect the decision boundary, as the classifier focuses on maximizing the margin around the support vectors \cite{zhu20031, scholkopf2002learning}.

\begin{figure}[hbt!]
\centering
\includegraphics[scale=0.291]{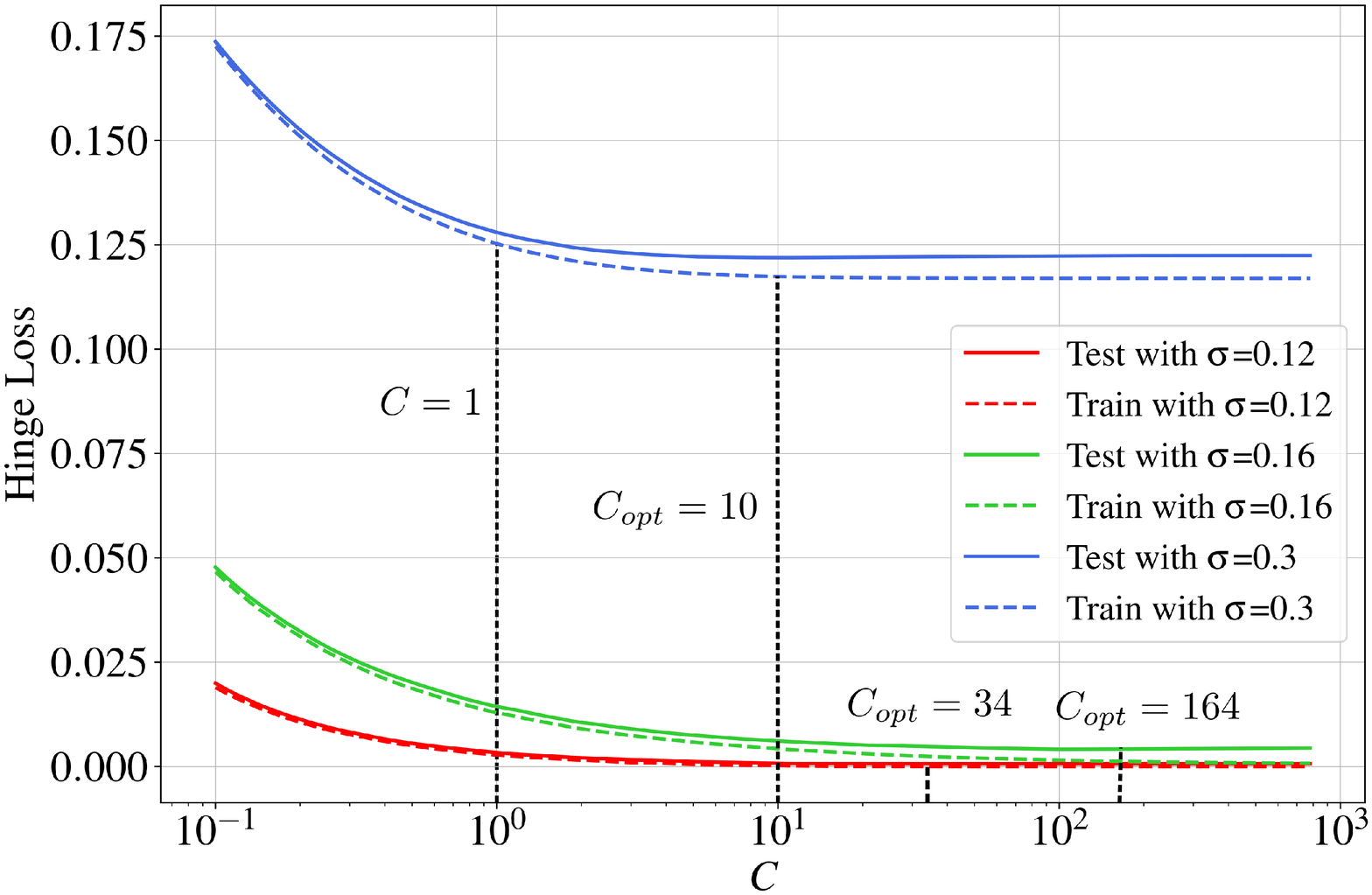}
\caption{Hinge loss for the train and test data set versus RP-$C$.}\label{fig:Hingloss_all}
\end{figure}

Figure \ref{fig:Hingloss_all} shows the hinge loss for the test and training data set in (\ref{eqn:Coptwithsigma12}) for three scenarios with $\sigma_1 = \sigma_2 = \sigma$ by averaging over 1000 runs.  The generated data sets were divided into a training dataset (70\%) and testing dataset(30\% ). As the figure indicates, with the gradual increase in $\sigma$ or decrease of S\&S values, due over lap, the hinge loss increases. Furthermore, the difference between the hinge loss of the training data set and the hinge loss of the test data set also increases. As the figure shows, the optimal value of RP-$C$, $C_{opt}$, which is selected with respect to the minimum hinge loss of the test data is $C_{opt}=34$ for $\sigma=0.12$, $C_{opt}=164$ for $\sigma=0.16$ and $C_{opt}=10$  for $\sigma=0.3$. It is worth mentioning that while all of these cases represent linearly separable classes due to the value of their $\sigma$, the optimal $C$ that minimized the hinge loss for all of them is very different from the default value $C=1$ for the linearly separable cases.

\begin{figure}[hbt!]
\centering
\includegraphics[scale=0.291]{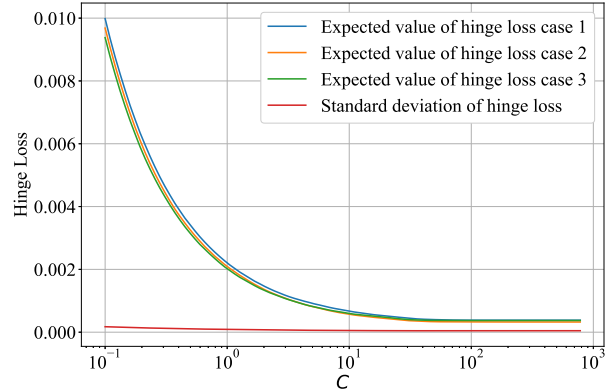}
\caption{Expected value and standard deviation of hinge loss for the train and test data set versus RP-$C$ for $\sigma=0.12$.}\label{fig:hing_sigmas}
\end{figure}

Figure \ref{fig:hing_sigmas} shows the behavior of the mean and standard deviation hinge loss with respect to RP-$C$. The expected behavior of the values is examined by averaging over 1000 runs. 
The results are shown for the three cases that are explained in the previous section and all have the same scatteredness $\sigma=0.12$. Note that the standard deviations of all cases are much smaller than the expected values. Furthermore, as the figure shows, as long as the scatteredness factor $\sigma$ remains the same, the hinge loss behavior exhibits a similar pattern and undergoes minimal variation. Consequently, this analysis indicates that the $C_{opt}$ look-up table  (\ref{eqn:Coptwithsigma12})that is function of  $d$ and $\sigma_1$ and $\sigma_2$, for the hinge loss is a function of $d$ and scatterness $\sigma$ in (\ref{eq:29}). Therefore, the scaled result is as follows.
\begin{align}
     E(L_H(C, S, \sigma_1, \sigma_2, d)) &=  \rho E(L_H(C, S, \sigma, \sigma, d))  \label{eqn:hingetaE} \\ &=  \lambda E(L_H(C, S, \frac{\sigma}{d}, \frac{\sigma}{d}, 1)) \label{eqn:higelamgda1}
\end{align}
where $\rho$ and $\lambda$ are some scaling functions that are not functions of $C$. Equivalently, $C_{opt}$ for hinge loss is function of S\&S ratio in (\ref{eq:30})

\begin{figure}[hbt!]
\begin{center}
\includegraphics[scale=0.27]{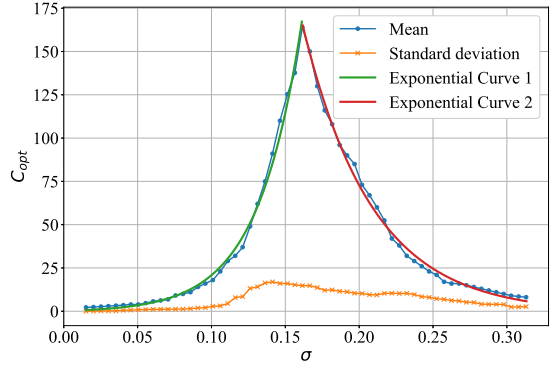}
\caption{Optimal regularization parameter as a function of $\frac{\sigma}{d}$ (d=1).}\label{fig:ORP_C_C}
\end{center}
\end{figure}

\begin{figure}[hbt!]
\centering
\includegraphics[scale=0.27]{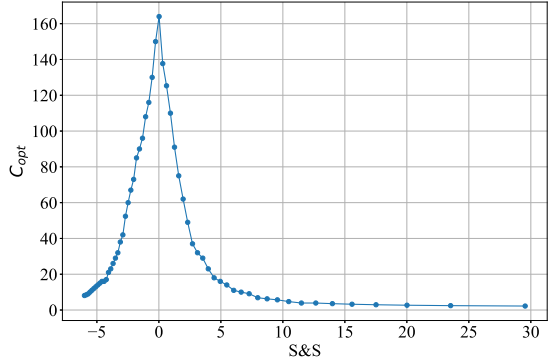}
\caption{Optimal regularization parameter and as a function of S\&S ratio }\label{fig:ORP_C_S}
\end{figure}

Figure \ref{fig:ORP_C_C} shows the calculated $C_{opt}$ based on the hinge loss for different values of scatterdness $\sigma$. Interestingly, the optimum value increases as a value of $\sigma$ up to $\sigma < 0.1666$ that is decreasing the margin.
However, as $\sigma$ grows or S\&S decreases, the probability of overlap between the data points of the two classes also increases and therefore a lower value of RP-$C$ is selected, which results in a trade-off between the SVM margin width and the generalization of the model to the test data set. For values of $\sigma$ larger than $0.3$, the probability of overlap between the two classes is very large, and the values of the calculated $C$ produce a very low accuracy.
Equivalently, Figure \ref{fig:ORP_C_S} shows the same selected $C_{opt}$ as a function of the  S\&S ratio. 

\begin{figure}[hbt!]
\begin{center}
\includegraphics[scale=0.5]{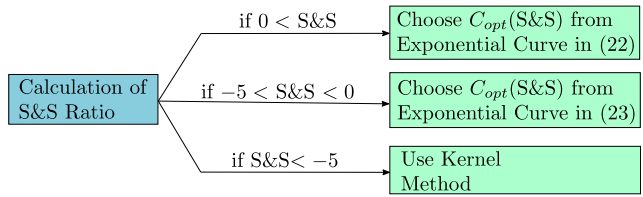}
\caption{Optimal RP-$C$ selection based on S\&S ratio.}\label{fig:SFCHART}
\end{center}
\end{figure}

Note that the breaking point $\sigma = 0.1666$ indicates that $S\&S= 0.0$. In addition, very interestingly, it seems that the behavior of the $C_{opt}$ can be well modeled with exponential curves as shown in Figure \ref{fig:ORP_C_C}, with the following equations:
\begin{align}
    C_{opt} (\mathrm{S\&S}) &= 0.7345 e^{(33.6915 \frac{\sigma}{d})} + -0.5247 \nonumber\\
    \mathrm{for}\;(\frac{\sigma}{d} &< 0.16)\label{eqn:C_i} \\
    C_{opt} (\mathrm{S\&S}) &= 5164.4657 e^{(-21.2514 \frac{\sigma}{d})} + -0.8548 \nonumber \\ \mathrm{for}\; (0.16&<\frac{\sigma}{d}<0.3)\label{eqn:C_d}
\end{align}
As $\sigma$ becomes larger than $0.3$, two classes are no longer linearly separable, and in this case, the SVM with kernel method should be implemented.
Corresponding values of S\&S ratio for scatterdness values $\sigma=0.1666$ and $\sigma=0.3$ are $0$ and $-5$. Therefore, the choice of $C_{opt}$ based on the S\&S ratio is as presented in Figure \ref{fig:SFCHART}.

\section{S\&S Ratio-Based Kernel Selection and Multi Class Classification}
\label{sec:Kernel}
 In many practical applications, there is no hyperplane that separates the two classes in the original low-dimensional space of the data points. In this case, every data point is mapped to a higher dimensional space, denoted by $F$, where it is possible to find a separating hyperplane. Consider $\phi$ as the mapping function or the transformation function such that $\phi: X \rightarrow F$ where $X$ denotes the original space of the data points. Consider the data points $\mathbf{x}_i, \mathbf{x}_j \in X$. The kernel function $K$ in terms of $\phi$ is expressed as follows:
\begin{align}
    K(\mathbf{x}_i, \mathbf{x}_j) = \phi(\mathbf{x}_i)^T \phi(\mathbf{x}_j)
\end{align}
 As in the SVM optimization the data points appear only in the form of inner products; it is possible to replace the inner product with a kernel function without any knowledge about the mapping function itself. 
 
 Starting with $n$ data points, the distance matrix obtained from the kernel function $K$, denoted by $\mathbf{G}$, is a $n \times n$ matrix of pairwise distances between any two data points in $F$ where the element $g_{ij} = g_{ji}$ denotes the distance between the $i^{\mathrm{th}}$ and $j^{\mathrm{th}}$ data points. Note that to ensure that $K$ is a kernel function, there must exist a feature space such as $F$ for which $K$ defines a dot product and the distance matrix $\mathbf{G}$ obtained from $K$ must be symmetric and a positive semidefinite matrix.  
 
 \subsection{S\&S Ratio-Based Optimal Kernel with Optimal Kernel Parameter Selection}
Conventional Grid search method finds the optimal kernel by defining a grid of kernel types with corresponding kernel parameters. Then the method loops through all possible combinations of the parameter values and calculates the performance score on the validation set. The best kernel with optimal parameters is chosen with respect to the maximum value of the performance score. Consider the set of possible kernels $\mathcal{K} = \{\kappa_1, \kappa_2, \dots, \kappa_h\}$, with a range of possible kernel parameters $\boldsymbol{\tau}_{\mathcal{K}}:\left [\tau_{\min}: \Delta_\tau: \tau_{\max}\right ]$, conventional grid search cross-validation method considers a grid of regularization parameters $C \in \left [C_{\min}: \Delta_C: C_{\max}\right ]$, $\kappa \in \mathcal{K}$, and $\tau \in \boldsymbol{\tau}_{\mathcal{K}}$ and determines the optimal hyperparameters as:
% \begin{align}
%     C_{opt} = \underset{C , \mathcal{K}, \boldsymbol{\tau}_{\mathcal{K}}}{\arg \min}L(C,\phi(S)) \label{eqn:coptcost}
% \end{align}
\begin{align}    (\mathrm{Grid\,Search\,CV})\,C_{opt} = \underset{\begin{matrix}
\mathcal{K} = \{K_1, \dots, K_h\}\\ 
\boldsymbol{\tau}_{\mathcal{K}}:\left [\tau_{\min}: \Delta_\tau: \tau_{\max}\right ]\\ 
C \in \left [C_{\min}: \Delta_C: C_{\max}\right ]
\end{matrix}}{\arg \min}L(C,\phi(X))
\end{align}
 
On the other hand, S\&S ratio is an indicator of separability and scatterdness and  can be used to compare kernels before using them in the SVM procedure. The kernel with maximum S\&S is the one that has separated the data the most. In order to calculate the S\&S ratio in kernel setting, the feature map of the data, $\phi(X)$, is used instead of $x$ itself in (\ref{eq:30}). The features can be found by calculating the SVD of the kernel matrix. However, kernel approximation methods are alternative approaches that use the kernel matrix to provide the feature map estimate. Using kernel approximate methods instead of SVD can be advantageous in specific cases, particularly for online learning and when working with very large datasets \cite{rahimi2007random}.

 There are different kernel approximation methods which provide a feature map from the kernel matrix. Random Kitchen Sinks (RKS) is a method to approximate shift-invariant kernel functions in machine learning, particularly in the context of large-scale learning problems \cite{rahimi2007random}. RKS method overcomes the computational challenges associated with kernel-based algorithms by constructing low-dimensional, randomized feature maps to approximate the original kernel function. This approach facilitates efficient learning on large datasets while preserving the benefits of kernel methods. The RKS method is primarily applied to shift-invariant kernels, such as the Radial Basis Function (RBF) kernel. 
 On the other hand, Polynomial Kernel Approximation via Tensor Sketch is a technique used to approximate polynomial kernel functions in machine learning. The Tensor Sketch method aims to reduce the computational cost associated with kernel-based learning algorithms by constructing a low-dimensional, randomized feature map that approximates the original polynomial kernel function \cite{pham2013fast}. Consequently, the approximation of $\phi(X)$ is calculated for the given kernel matrix.
The resulted $\phi(X)$ from the kernel approximation approach produces $d_\kappa$ and $\sigma_\kappa$. Denote the S\&S of each kernel $\kappa$ and with its parameter $\tau$ as $\mathrm{S\&S}_{\kappa, \tau}$. Note that the only acceptable kernels based on the S\&S are the ones with S\&S$> -5$, and the method is able to eliminate the unsuitable kernels at this stage before using the SVM. 
Therefore, the best kernel to be used in the SVM procedure is the one with the highest value of S\&S: 
\begin{align}
    (\kappa_{opt}, \tau_{opt}) = \underset{\kappa \in \mathcal{K}, \tau \in \boldsymbol{\tau}}{\arg \max} \, (\mathrm{S}\&\mathrm{S} _{\kappa, \tau}) \label{eqn:coptcostkernel} \\
    \mathrm{S\&S}_{opt} = \mathrm{S}\&\mathrm{S} _{\kappa_{opt}, \tau_{opt}}
\end{align}

which leads to the following optimization for optimal RP-$C$:
\begin{align}
    C_{opt} = \left\{\begin{matrix}
    C_{opt}(\mathrm{S}\&\mathrm{S} _{\kappa_{opt}, \tau_{opt}})\, \mathrm{from (\ref{eqn:C_i})} & \mathrm{if} &\mathrm{S\&S} > 0 \\  
    C_{opt}(\mathrm{S}\&\mathrm{S} _{\kappa_{opt}, \tau_{opt}})\, \mathrm{from (\ref{eqn:C_d})} & \mathrm{if} & -5<\mathrm{S\&S} < 0\\  
\end{matrix}\right. \label{eqn:coptkrl}
\end{align}

\subsection{Complexity Analysis}
\label{sec:complexity}
\textit{Computational complexity of grid search CV:} The computational complexity of grid search cross-validation for Support Vector Machines (SVM) with a kernel method, while accounting for the kernel and regularization parameters is as follows:
\begin{equation}
 m_q m_C m_k \times O(n^3\psi)
\end{equation}
where $O(n^3\psi)$ is the computational complexity of the SVM model with $\psi$ the number of features, $m_k$ corresponds to the k-fold cross-validation, $m_q$ refers to the number of distinct values for the combinations of kernels and kernel parameters, and $m_C$ is the number of considered regularization parameters \cite{hsu2003practical, hastie2004entire, chang2011libsvm}. 

\textit{Computational complexity of the proposed method:} The complexity of the proposed method is due to the implementation of SVM once the S\&S is calculated. There is also the computational complexity of the kernel approximation methods to find the S\&S first.  The computational complexities of the two most popular kernel approximation methods, RKS and Tensor Sketch, are $O(D n)$ and $\mathcal{O}(n\psi+ D \log(D)))$  respectively, where $D$ is the dimensionality of the approximate feature map \cite{rahimi2007random} \cite{pham2013fast}. Therefore, for example, the computational complexity of the proposed method when RKS is used is:
\begin{align}
     m_q m_C O(n D) + O(n^3\phi)
\end{align}

\begin{figure}[!htb]
\centering
\begin{subfigure}{0.41\textwidth}
    \centering
    \includegraphics[width=\textwidth]{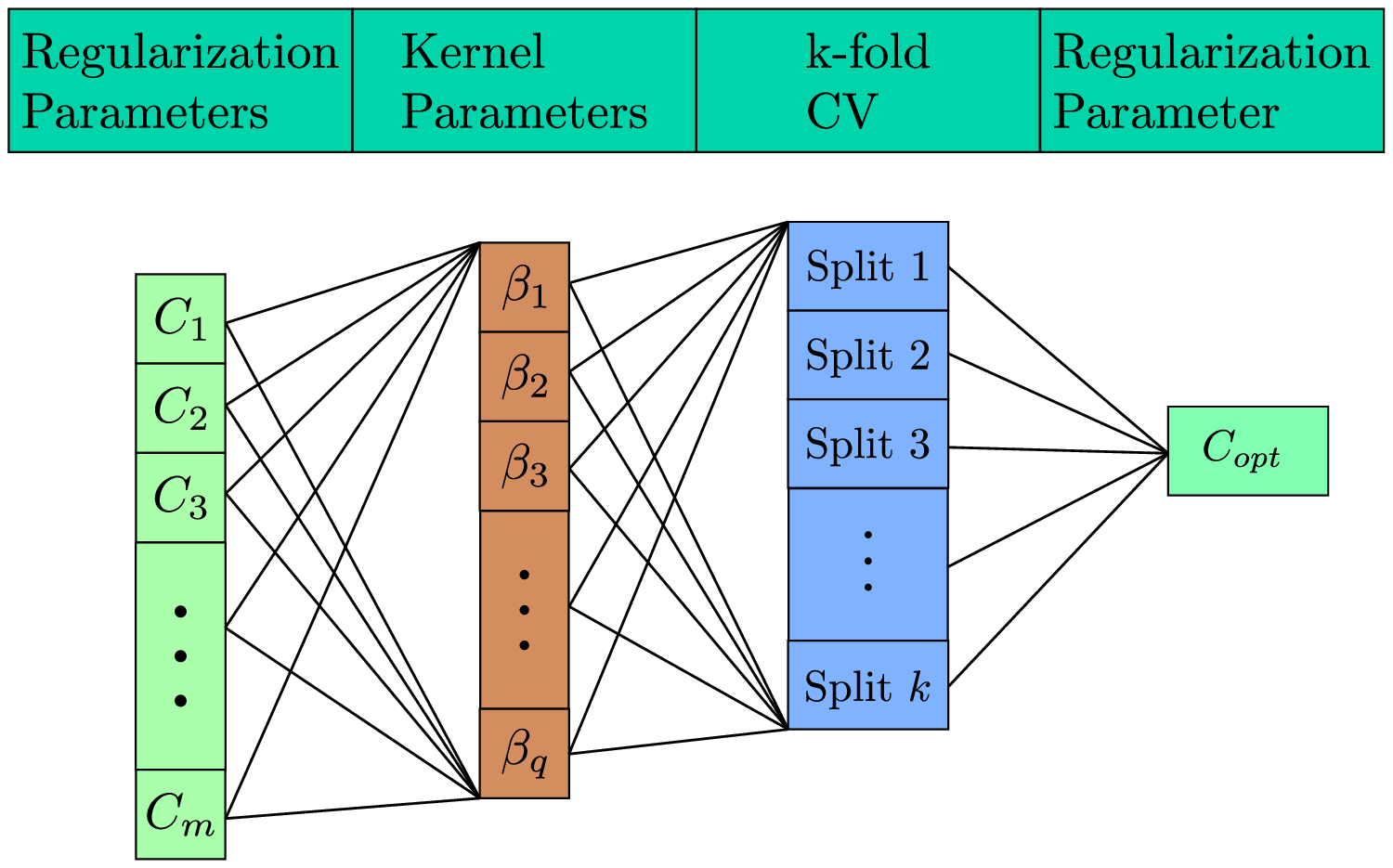}
    \caption{Grid Search with k-fold Cross-Validation}
    \label{fig:computCV}
\end{subfigure}
\hfill
% \vspace{0.5cm}
\begin{subfigure}{0.41\textwidth}
    \centering
    \includegraphics[width=\textwidth]{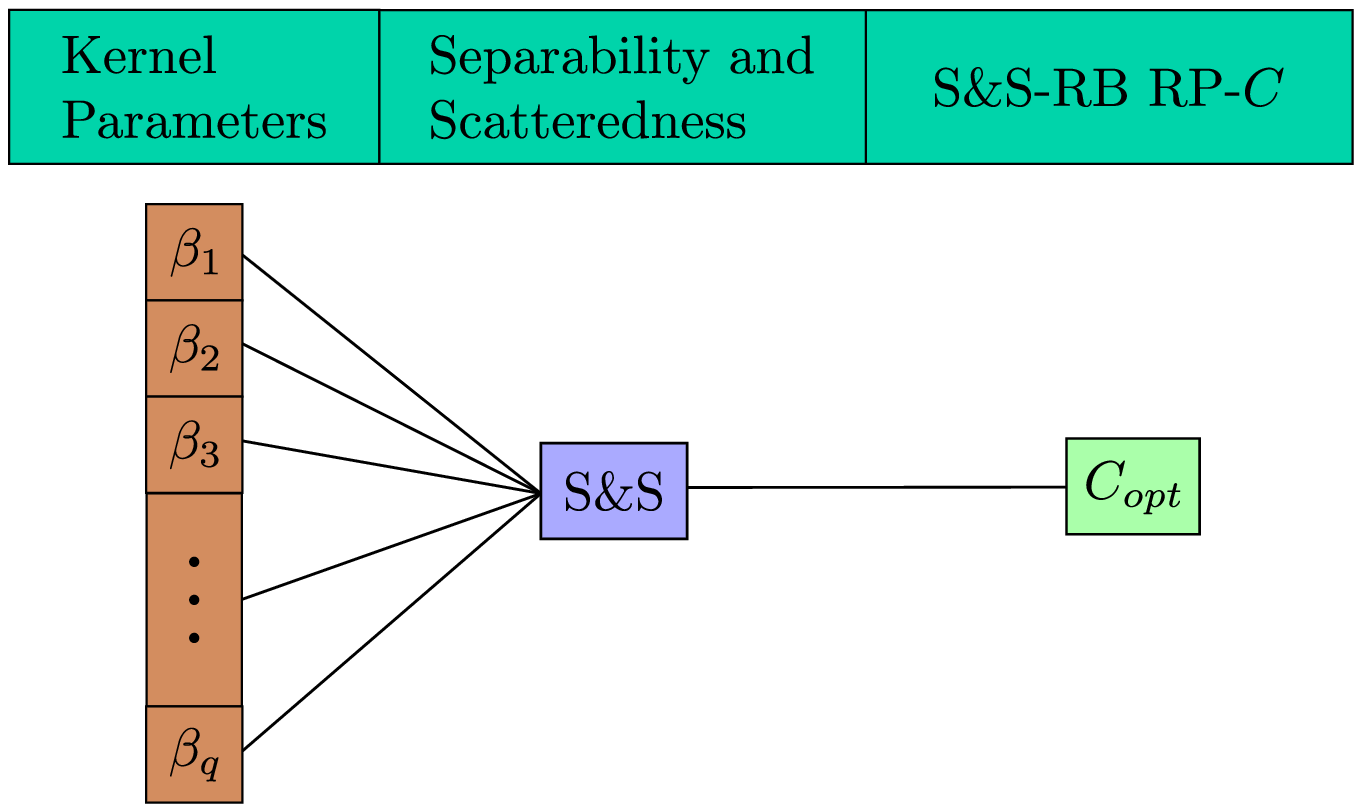}
    \caption{S\&S-RB}
    \label{fig:computsb}
\end{subfigure}
\hfill
\caption{Computation Complexity of the grid search with k-fold cross validation (\ref{fig:computCV}) and the proposed S\&S-RB method (\ref{fig:computsb}).}
\label{fig:comput}
\end{figure}

Figure \ref{fig:comput} visualizes the computational complexity of the grid search as well as the S\&S ratio-based approach for SVM with the kernel method. 

\subsection{Multi Class SVM Algorithm}  
\begin{figure}[hbt!]
\begin{center}
\includegraphics[scale=0.45]{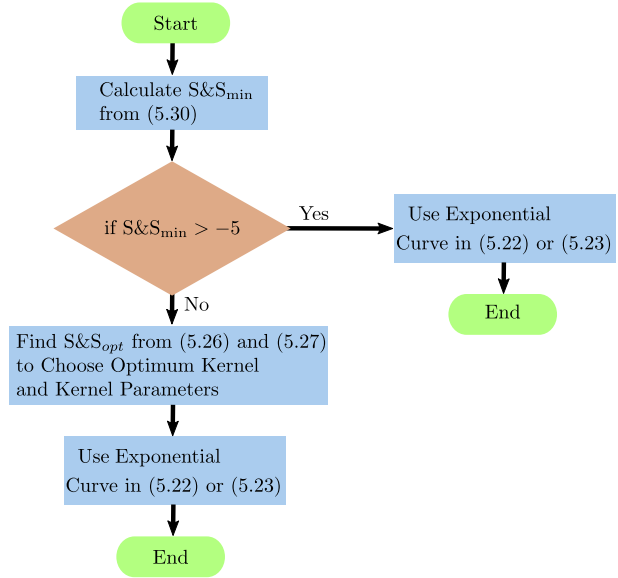}
\caption{The flow chart of the proposed S\&S-RB method for multiclass kernel SVM to determine the optimal regularization parameter $C$.}\label{fig:FLCH2}
\end{center}
\end{figure}

For multiclass SVM algorithm, the most common approaches are One-vs-One (OvO) and One-vs-All (OvA, also called One-vs-Rest). Both of these methods involve breaking down the multiclass problem into multiple binary classification problems and then combining the results. Therefore, if the total number of classes in a multiclass SVM is $r$, the grid-search cross-validation method must be run $\binom{r}{2}$ times for each pair of classes. In the proposed S\&S based method, first the kernel approximation is calculatd once for the input data set and consequently, the distance $d$, scatteredness $\sigma$ and S\&S values are calculated for each pair of classes. The S\&S for class $i$ and class $j$ is denoted as $\mathrm{S}\&\mathrm{S}_{(i, j)}$ and the lowest calculated S\&S among pairs of classes is:
\begin{align}
    \mathrm{S}\&\mathrm{S}_{\min} = \underset{i, j \in r}{\min} \, \mathrm{S}\&\mathrm{S}_{(i, j)} \label{eqn:SNLow}
\end{align}
This lowest calculated $\mathrm{S}\&\mathrm{S}_{\min}$ is the worst-case S\&S and can be used to determine the optimum RP-$C$ as:
\begin{align}
    C_{opt} = \left\{\begin{matrix}
    C_{opt}(\mathrm{S\&S\,}_{\min})\, \mathrm{from (\ref{eqn:C_i})} & \mathrm{if} &\mathrm{S\&S} > 0 \\  
    C_{opt}(\mathrm{S\&S\,}_{\min})\, \mathrm{from (\ref{eqn:C_d})} & \mathrm{if} & -5<\mathrm{S\&S} < 0\\  
    C_{opt}(\mathrm{S\&S\,}_{\min})\, \mathrm{from (\ref{eqn:coptkrl})}  & \mathrm{if} & \mathrm{S\&S} < -5 
\end{matrix}\right. \label{eqn:coptSandSij}
\end{align}
In multiclass SVM employing kernel methods, the S\&S values are computed for each pair of classes utilizing various kernel functions with a range of kernel parameters. Subsequently, the minimum S\&S value among the class pairs is designated as the representative S\&S value for each kernel with specific parameters. Figure \ref{fig:FLCH2} shows the flow chart of the proposed S\&S-RB method for multi class SVM with kernel method.

\section{Simulation Results}
\label{section:simu}
In this section, the performance of the proposed S\&S-RB method is evaluated and compared with the grid search 5-fold cross-validation technique using real datasets from the Keel repository \cite{alcala2009keel}, the UCI data repository \cite{Dua:2019}, and the MNIST \cite{deng2012mnist} datasets. Table \ref{tab:BVBSIm} presents the descriptions of the data sets. Three distinct kernel functions are employed, the Radial Basis Function (RBF), the polynomial, and the sigmoid, to transform the data set into a higher-dimensional space. Table \ref{tab:BVBSIm} also shows the calculated S$\&\mathrm{S}_{\min}$ according to (\ref{eqn:SNLow}) for each dataset, as well as the corresponding distance ($d$) and the scatteredness factor ($\sigma$).
%------------------------------------ table dataset information ------------------------------------------
% \usepackage{graphicx}
\begin{table*}[]
\centering
\caption{Description of the datasets with the lowest calculated value for S$\&\mathrm{S}_{\min}$}
\label{tab:BVBSIm}
% \resizebox{\textwidth}{!}{%
\begin{tabular}{|l|l|l|l|l|l|l|}
\hline
Dataset                 & Size  & \# Feature & \# Class & $d$   & $\sigma$ & S\&S \\ \hline
eeg-eye-state           & 14980 & 15         & 2        & 5.87  & 1.1336   & -1.28  \\ \hline
spambase                & 4601  & 58         & 2        & 5.41  & 0.8045   & 0.99   \\ \hline
mnist\_784               & 70000 & 785        & 10       & 1.38  & 0.2124   & 0.69   \\ \hline
waveform-5000           & 5000  & 41         & 3        & 4.23  & 0.8235   & -1.35  \\ \hline
adult                   & 48842 & 15         & 2        & 2.16  & 0.4464   & -1.87  \\ \hline
appendicitis            & 106   & 7          & 2        & 3.28  & 0.7537   & -2.79  \\ \hline
australian              & 690   & 14         & 2        & 3.45  & 0.5089   & 1.06   \\ \hline
balance                 & 525   & 4          & 3        & -2.59 & 0.9746   & -2.59  \\ \hline
banana                  & 5302  & 2          & 2        & -0.88 & 0.1604   & -0.88  \\ \hline
bands                   & 365   & 19         & 2        & 2.44  & 0.375    & 2.44   \\ \hline
banknote authentication & 1372  & 4          & 2        & 4.78  & 0.695    & 4.78   \\ \hline
breast tissue           & 106   & 9          & 4        & 1.45  & 0.3808   & -3.95  \\ \hline
haberman                & 306   & 3          & 2        & 1.98  & 0.4571   & -2.83  \\ \hline
hayes-roth              & 160   & 4          & 3        & 1.98  & 0.4036   & -1.75  \\ \hline
heart                   & 270   & 13         & 2        & 1.87  & 0.3529   & -1.08  \\ \hline
hepatitis               & 80    & 19         & 2        & 2.19  & 0.3264   & 0.97   \\ \hline
ilpd                    & 583   & 9          & 2        & 3.85  & 0.6604   & -0.25  \\ \hline
iris                    & 150   & 4          & 3        & 4.54  & 0.5369   & 2.98   \\ \hline
mammographic            & 830   & 5          & 2        & 1.56  & 0.3531   & -2.66  \\ \hline
newthyroid              & 215   & 5          & 3        & 1.78  & 0.2361   & 1.98   \\ \hline
penbased\_1(5,8,9)      & 3165  & 16         & 3        & 1.56  & 0.3479   & -2.53  \\ \hline
tae                     & 151   & 5          & 3        & 2.89  & 0.6512   & -2.62  \\ \hline
pima                    & 769   & 8          & 2        & 3.97  & 0.8339   & -2.01  \\ \hline
\end{tabular}%
% }
\end{table*}

\begin{figure}[hbt!]
\begin{center}
\includegraphics[scale=0.29]{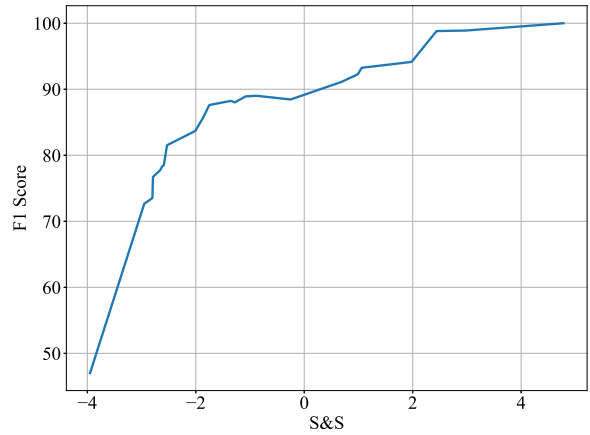}
\caption{F1 accuracy score versus S\&S value.}\label{fig:SandSF1}
\end{center}
\end{figure}

Table \ref{tab:simulation_final} compares the SVM algorithm with the regularization parameter selected from S\&S ratio-based (S\&S-RB) method and with the grid search 5-fold cross-validation. For the cross validation approach, CV (FI) and CV( Hinge) are the results of the method with F1 and hinge loss scores respectively. As indicated in the table, the regularization parameter chosen by the proposed S\&S-RB method has a performance similar to or even better than grid search cross-validation; however, the cross-validation method requires much more computational time compared to S\&S-RB approach which is due to the computational complexity of the approaches as discussed in Section \ref{sec:complexity}. The grid search cross-validation method requires comparing all possible combinations of kernels, kernel parameters, and RP-$C$ which results in more computational complexity; whereas in the proposed method, the optimal kernel and kernel parameters can be obtained using the method in Section \ref{sec:Kernel} and the S\&S based optimization in (\ref{eqn:SNLow}) before the SVM is implemented. 

Figure \ref{fig:SandSF1} shows the F1 score of the proposed method versus S\&S values in Tables \ref{tab:BVBSIm} and \ref{tab:simulation_final}. As the figure shows, the F1 score of the SVM algorithm is correlated with the calculated S\&S values. The higher S\&S value indicate lower overlap between the data points in the two classes; therefore, the separating hyperplane of the SVM model can generalize better to the test data set. As a result, the calculated S\$S values from the kernel method are perfect indicators of the performance of the SVM model before starting the process, i.e., similar to Signal to Noise ratio (SNR) that can be used to predict the performance of a system, the S\&S can be used before the SVM procedure to evaluate or compare the accuracy of the results.

\begin{table*}[]
\centering
\caption{Simulation results in terms of the F1 score and the run time for Cross-validation with Cross-Validation (with two cost functions, hinge loss) and S\&S-RB method.}
\label{tab:simulation_final}
\resizebox{\textwidth}{!}{%
\begin{tabular}{|l|lll|lll|}
\hline
                        & \multicolumn{3}{l|}{F1-Score}                                              & \multicolumn{3}{l|}{Run Time (Sec)}                                        \\ \hline
Dataset                 & \multicolumn{1}{l|}{CV (F1)} & \multicolumn{1}{l|}{CV (Hinge)} & S\&S-RB & \multicolumn{1}{l|}{CV (F1)} & \multicolumn{1}{l|}{CV (Hinge)} & S\&S-RB \\ \hline
eeg-eye-state           & \multicolumn{1}{l|}{87.98}   & \multicolumn{1}{l|}{88.01}      & 88.01     & \multicolumn{1}{l|}{832.059} & \multicolumn{1}{l|}{865.34}     & 23.14     \\ \hline
spambase                & \multicolumn{1}{l|}{91.77}   & \multicolumn{1}{l|}{92.03}      & 92.29     & \multicolumn{1}{l|}{128.24}  & \multicolumn{1}{l|}{130.41}     & 6.28      \\ \hline
mnist\_784                   & \multicolumn{1}{l|}{91.08}   & \multicolumn{1}{l|}{91.08}      & 91.12     & \multicolumn{1}{l|}{3452.14} & \multicolumn{1}{l|}{3677.28}    & 124.14    \\ \hline
waveform-5000           & \multicolumn{1}{l|}{86.74}   & \multicolumn{1}{l|}{86.74}      & 88.24     & \multicolumn{1}{l|}{200.45}  & \multicolumn{1}{l|}{210.53}     & 9.81      \\ \hline
adult                   & \multicolumn{1}{l|}{82.41}   & \multicolumn{1}{l|}{84.44}      & 85.62     & \multicolumn{1}{l|}{1255.14} & \multicolumn{1}{l|}{1156.52}    & 15.67     \\ \hline
appendicitis            & \multicolumn{1}{l|}{74.69}   & \multicolumn{1}{l|}{76.71}      & 76.71     & \multicolumn{1}{l|}{73.41}   & \multicolumn{1}{l|}{73.24}      & 1.05      \\ \hline
australian              & \multicolumn{1}{l|}{91.81}   & \multicolumn{1}{l|}{92.74}      & 93.25     & \multicolumn{1}{l|}{63.32}   & \multicolumn{1}{l|}{62.42}      & 0.71      \\ \hline
balance                 & \multicolumn{1}{l|}{76.65}   & \multicolumn{1}{l|}{78.48}      & 78.48     & \multicolumn{1}{l|}{45.57}   & \multicolumn{1}{l|}{44.44}      & 0.03      \\ \hline
banana                  & \multicolumn{1}{l|}{89.01}   & \multicolumn{1}{l|}{89.01}      & 89.01     & \multicolumn{1}{l|}{61.33}   & \multicolumn{1}{l|}{63.15}      & 0.63      \\ \hline
bands                   & \multicolumn{1}{l|}{98.22}   & \multicolumn{1}{l|}{98.81}      & 98.81     & \multicolumn{1}{l|}{75.51}   & \multicolumn{1}{l|}{74.62}      & 0.97      \\ \hline
banknote authentication & \multicolumn{1}{l|}{100}     & \multicolumn{1}{l|}{100}        & 100       & \multicolumn{1}{l|}{10.17}   & \multicolumn{1}{l|}{11.04}      & 0.01      \\ \hline
breast tissue           & \multicolumn{1}{l|}{41.44}   & \multicolumn{1}{l|}{45.51}      & 47.01     & \multicolumn{1}{l|}{15.02}   & \multicolumn{1}{l|}{15.23}      & 0.01      \\ \hline
diabetes                & \multicolumn{1}{l|}{72.62}   & \multicolumn{1}{l|}{72.55}      & 72.66     & \multicolumn{1}{l|}{80.12}   & \multicolumn{1}{l|}{80.41}      & 0.94      \\ \hline
haberman                & \multicolumn{1}{l|}{72.24}   & \multicolumn{1}{l|}{73.52}      & 73.52     & \multicolumn{1}{l|}{44.23}   & \multicolumn{1}{l|}{44.58}      & 0.07      \\ \hline
hayes-roth              & \multicolumn{1}{l|}{86.42}   & \multicolumn{1}{l|}{87.14}      & 87.61     & \multicolumn{1}{l|}{7.12}    & \multicolumn{1}{l|}{7.11}       & 0.01      \\ \hline
heart                   & \multicolumn{1}{l|}{88.62}   & \multicolumn{1}{l|}{88.52}      & 88.92     & \multicolumn{1}{l|}{12.14}   & \multicolumn{1}{l|}{12.01}      & 0.01      \\ \hline
hepatitis               & \multicolumn{1}{l|}{90.3}    & \multicolumn{1}{l|}{92.2}       & 92.2      & \multicolumn{1}{l|}{16.62}   & \multicolumn{1}{l|}{16.72}      & 3.08      \\ \hline
ilpd                    & \multicolumn{1}{l|}{86.31}   & \multicolumn{1}{l|}{87.35}      & 88.45     & \multicolumn{1}{l|}{7.56}    & \multicolumn{1}{l|}{7.27}       & 0.41      \\ \hline
iris                    & \multicolumn{1}{l|}{98.23}   & \multicolumn{1}{l|}{98.87}      & 98.89     & \multicolumn{1}{l|}{1.24}    & \multicolumn{1}{l|}{1.14}       & 0.04      \\ \hline
mammographic            & \multicolumn{1}{l|}{78.15}   & \multicolumn{1}{l|}{77.63}      & 77.71     & \multicolumn{1}{l|}{10.51}   & \multicolumn{1}{l|}{10.17}      & 0.01      \\ \hline
newthyroid              & \multicolumn{1}{l|}{93.05}   & \multicolumn{1}{l|}{94.14}      & 94.14     & \multicolumn{1}{l|}{4.86}    & \multicolumn{1}{l|}{4.75}       & 0.01      \\ \hline
penbased\_1(5,8,9)      & \multicolumn{1}{l|}{81.24}   & \multicolumn{1}{l|}{81.52}      & 81.52     & \multicolumn{1}{l|}{77.52}   & \multicolumn{1}{l|}{78.02}      & 1.13      \\ \hline
tae                     & \multicolumn{1}{l|}{77.34}   & \multicolumn{1}{l|}{77.21}      & 78.31     & \multicolumn{1}{l|}{3.41}    & \multicolumn{1}{l|}{3.22}       & 0.01      \\ \hline
pima                    & \multicolumn{1}{l|}{84.67}   & \multicolumn{1}{l|}{83.57}      & 83.68     & \multicolumn{1}{l|}{4.08}    & \multicolumn{1}{l|}{4.11}       & 0.01      \\ \hline
\end{tabular}%
}
\end{table*}

\section{Conclusion}
An alternative approach, compared to conventional grid search cross-validation, is proposed for hyperparameter selection in the SVM algorithm. Grid-search cross-validation is a systematic approach that evaluates model performance across different combinations of hyperparameters to identify the best set of regularization parameter (RP), kernel type, and kernel parameters. This search can be computationally expensive and exhaustive, in particular when dealing with high-dimensional data as it has to compare the performance of the SVM for each element of the hyperparameter grid. The proposed Separability and Scatteredness (S\&S) ratio-based approach provides the optimum value of the hyperparameter using a lookup table and therefore eliminates the need for a exhaustive grid search. 
The approach offers a computationally efficient alternative to traditional grid search cross-validation techniques while maintaining or even surpassing their performance in terms of accuracy. 
Furthermore, S\&S ratio is a reliable indicator in predicting the performance of the SVM model, with higher values indicating lower overlap between data points in different classes and, consequently, better generalization capabilities. It was shown that the S\&S  ratio can be used as a valuable metric to identify non-separable datasets that require the kernel method. The calculated S\&S ratio not only enables the learner to choose the optimum kernel and kernel parameters from the lookup table and without the need to search in a grid, but also can evaluate the performance of SVM model a priori. The role of the S\&S in SVM algorithms is analogous to the signal-to-noise ratio (SNR) used in signal processing or communication. The analysis started with the two-class SVM problem and was successfully extended to handle multi-class SVM scenarios. The simulation results on real data sets show that the proposed S\&S-RB method offers significant advantages in terms of computational efficiency, with similar or in occasions even better accuracy than the cross-validation grid-search.

\bibliography{sn-bibliography}% common bib file


\begin{thebibliography}{31}
\ifx \bisbn   \undefined \def \bisbn  #1{ISBN #1}\fi
\ifx \binits  \undefined \def \binits#1{#1}\fi
\ifx \bauthor  \undefined \def \bauthor#1{#1}\fi
\ifx \batitle  \undefined \def \batitle#1{#1}\fi
\ifx \bjtitle  \undefined \def \bjtitle#1{#1}\fi
\ifx \bvolume  \undefined \def \bvolume#1{\textbf{#1}}\fi
\ifx \byear  \undefined \def \byear#1{#1}\fi
\ifx \bissue  \undefined \def \bissue#1{#1}\fi
\ifx \bfpage  \undefined \def \bfpage#1{#1}\fi
\ifx \blpage  \undefined \def \blpage #1{#1}\fi
\ifx \burl  \undefined \def \burl#1{\textsf{#1}}\fi
\ifx \doiurl  \undefined \def \doiurl#1{\url{https://doi.org/#1}}\fi
\ifx \betal  \undefined \def \betal{\textit{et al.}}\fi
\ifx \binstitute  \undefined \def \binstitute#1{#1}\fi
\ifx \binstitutionaled  \undefined \def \binstitutionaled#1{#1}\fi
\ifx \bctitle  \undefined \def \bctitle#1{#1}\fi
\ifx \beditor  \undefined \def \beditor#1{#1}\fi
\ifx \bpublisher  \undefined \def \bpublisher#1{#1}\fi
\ifx \bbtitle  \undefined \def \bbtitle#1{#1}\fi
\ifx \bedition  \undefined \def \bedition#1{#1}\fi
\ifx \bseriesno  \undefined \def \bseriesno#1{#1}\fi
\ifx \blocation  \undefined \def \blocation#1{#1}\fi
\ifx \bsertitle  \undefined \def \bsertitle#1{#1}\fi
\ifx \bsnm \undefined \def \bsnm#1{#1}\fi
\ifx \bsuffix \undefined \def \bsuffix#1{#1}\fi
\ifx \bparticle \undefined \def \bparticle#1{#1}\fi
\ifx \barticle \undefined \def \barticle#1{#1}\fi
\bibcommenthead
\ifx \bconfdate \undefined \def \bconfdate #1{#1}\fi
\ifx \botherref \undefined \def \botherref #1{#1}\fi
\ifx \url \undefined \def \url#1{\textsf{#1}}\fi
\ifx \bchapter \undefined \def \bchapter#1{#1}\fi
\ifx \bbook \undefined \def \bbook#1{#1}\fi
\ifx \bcomment \undefined \def \bcomment#1{#1}\fi
\ifx \oauthor \undefined \def \oauthor#1{#1}\fi
\ifx \citeauthoryear \undefined \def \citeauthoryear#1{#1}\fi
\ifx \endbibitem  \undefined \def \endbibitem {}\fi
\ifx \bconflocation  \undefined \def \bconflocation#1{#1}\fi
\ifx \arxivurl  \undefined \def \arxivurl#1{\textsf{#1}}\fi
\csname PreBibitemsHook\endcsname

%%% 1
\bibitem[\protect\citeauthoryear{Gao et~al.}{2002}]{gao2002probabilistic}
\begin{barticle}
\bauthor{\bsnm{Gao}, \binits{J.B.}},
\bauthor{\bsnm{Gunn}, \binits{S.R.}},
\bauthor{\bsnm{Harris}, \binits{C.J.}},
\bauthor{\bsnm{Brown}, \binits{M.}}:
\batitle{A probabilistic framework for svm regression and error bar
  estimation}.
\bjtitle{Machine Learning}
\bvolume{46},
\bfpage{71}--\blpage{89}
(\byear{2002})
\end{barticle}
\endbibitem

%%% 2
\bibitem[\protect\citeauthoryear{Lenz et~al.}{2022}]{lenz2022optimised}
\begin{barticle}
\bauthor{\bsnm{Lenz}, \binits{O.U.}},
\bauthor{\bsnm{Peralta}, \binits{D.}},
\bauthor{\bsnm{Cornelis}, \binits{C.}}:
\batitle{Optimised one-class classification performance}.
\bjtitle{Machine Learning}
\bvolume{111}(\bissue{8}),
\bfpage{2863}--\blpage{2883}
(\byear{2022})
\end{barticle}
\endbibitem

%%% 3
\bibitem[\protect\citeauthoryear{Widodo and Yang}{2007}]{widodo2007support}
\begin{barticle}
\bauthor{\bsnm{Widodo}, \binits{A.}},
\bauthor{\bsnm{Yang}, \binits{B.-S.}}:
\batitle{Support vector machine in machine condition monitoring and fault
  diagnosis}.
\bjtitle{Mechanical systems and signal processing}
\bvolume{21}(\bissue{6}),
\bfpage{2560}--\blpage{2574}
(\byear{2007})
\end{barticle}
\endbibitem

%%% 4
\bibitem[\protect\citeauthoryear{Byvatov and
  Schneider}{2003}]{byvatov2003support}
\begin{barticle}
\bauthor{\bsnm{Byvatov}, \binits{E.}},
\bauthor{\bsnm{Schneider}, \binits{G.}}:
\batitle{Support vector machine applications in bioinformatics.}
\bjtitle{Applied bioinformatics}
\bvolume{2}(\bissue{2}),
\bfpage{67}--\blpage{77}
(\byear{2003})
\end{barticle}
\endbibitem

%%% 5
\bibitem[\protect\citeauthoryear{Dino and Abdulrazzaq}{2019}]{dino2019facial}
\begin{bchapter}
\bauthor{\bsnm{Dino}, \binits{H.I.}},
\bauthor{\bsnm{Abdulrazzaq}, \binits{M.B.}}:
\bctitle{Facial expression classification based on svm, knn and mlp
  classifiers}.
In: \bbtitle{2019 International Conference on Advanced Science and Engineering
  (ICOASE)},
pp. \bfpage{70}--\blpage{75}
(\byear{2019}).
\bcomment{IEEE}
\end{bchapter}
\endbibitem

%%% 6
\bibitem[\protect\citeauthoryear{Kumar et~al.}{2019}]{kumar2019face}
\begin{barticle}
\bauthor{\bsnm{Kumar}, \binits{A.}},
\bauthor{\bsnm{Kaur}, \binits{A.}},
\bauthor{\bsnm{Kumar}, \binits{M.}}:
\batitle{Face detection techniques: a review}.
\bjtitle{Artificial Intelligence Review}
\bvolume{52}(\bissue{2}),
\bfpage{927}--\blpage{948}
(\byear{2019})
\end{barticle}
\endbibitem

%%% 7
\bibitem[\protect\citeauthoryear{Daberdaku and
  Ferrari}{2019}]{daberdaku2019antibody}
\begin{barticle}
\bauthor{\bsnm{Daberdaku}, \binits{S.}},
\bauthor{\bsnm{Ferrari}, \binits{C.}}:
\batitle{Antibody interface prediction with 3d zernike descriptors and svm}.
\bjtitle{Bioinformatics}
\bvolume{35}(\bissue{11}),
\bfpage{1870}--\blpage{1876}
(\byear{2019})
\end{barticle}
\endbibitem

%%% 8
\bibitem[\protect\citeauthoryear{Sanz et~al.}{2018}]{sanz2018svm}
\begin{barticle}
\bauthor{\bsnm{Sanz}, \binits{H.}},
\bauthor{\bsnm{Valim}, \binits{C.}},
\bauthor{\bsnm{Vegas}, \binits{E.}},
\bauthor{\bsnm{Oller}, \binits{J.M.}},
\bauthor{\bsnm{Reverter}, \binits{F.}}:
\batitle{Svm-rfe: selection and visualization of the most relevant features
  through non-linear kernels}.
\bjtitle{BMC bioinformatics}
\bvolume{19}(\bissue{1}),
\bfpage{1}--\blpage{18}
(\byear{2018})
\end{barticle}
\endbibitem

%%% 9
\bibitem[\protect\citeauthoryear{Dhar et~al.}{2021}]{dhar2021text}
\begin{barticle}
\bauthor{\bsnm{Dhar}, \binits{A.}},
\bauthor{\bsnm{Mukherjee}, \binits{H.}},
\bauthor{\bsnm{Dash}, \binits{N.S.}},
\bauthor{\bsnm{Roy}, \binits{K.}}:
\batitle{Text categorization: past and present}.
\bjtitle{Artificial Intelligence Review}
\bvolume{54}(\bissue{4}),
\bfpage{3007}--\blpage{3054}
(\byear{2021})
\end{barticle}
\endbibitem

%%% 10
\bibitem[\protect\citeauthoryear{Berge et~al.}{2019}]{berge2019using}
\begin{barticle}
\bauthor{\bsnm{Berge}, \binits{G.T.}},
\bauthor{\bsnm{Granmo}, \binits{O.-C.}},
\bauthor{\bsnm{Tveit}, \binits{T.O.}},
\bauthor{\bsnm{Goodwin}, \binits{M.}},
\bauthor{\bsnm{Jiao}, \binits{L.}},
\bauthor{\bsnm{Matheussen}, \binits{B.V.}}:
\batitle{Using the tsetlin machine to learn human-interpretable rules for
  high-accuracy text categorization with medical applications}.
\bjtitle{IEEE Access}
\bvolume{7},
\bfpage{115134}--\blpage{115146}
(\byear{2019})
\end{barticle}
\endbibitem

%%% 11
\bibitem[\protect\citeauthoryear{Blumer et~al.}{1989}]{blumer1989learnability}
\begin{barticle}
\bauthor{\bsnm{Blumer}, \binits{A.}},
\bauthor{\bsnm{Ehrenfeucht}, \binits{A.}},
\bauthor{\bsnm{Haussler}, \binits{D.}},
\bauthor{\bsnm{Warmuth}, \binits{M.K.}}:
\batitle{Learnability and the vapnik-chervonenkis dimension}.
\bjtitle{Journal of the ACM (JACM)}
\bvolume{36}(\bissue{4}),
\bfpage{929}--\blpage{965}
(\byear{1989})
\end{barticle}
\endbibitem

%%% 12
\bibitem[\protect\citeauthoryear{Suykens and
  Vandewalle}{1999}]{suykens1999least}
\begin{barticle}
\bauthor{\bsnm{Suykens}, \binits{J.A.}},
\bauthor{\bsnm{Vandewalle}, \binits{J.}}:
\batitle{Least squares support vector machine classifiers}.
\bjtitle{Neural processing letters}
\bvolume{9}(\bissue{3}),
\bfpage{293}--\blpage{300}
(\byear{1999})
\end{barticle}
\endbibitem

%%% 13
\bibitem[\protect\citeauthoryear{Wang}{2005}]{wang2005support}
\begin{bbook}
\bauthor{\bsnm{Wang}, \binits{L.}}:
\bbtitle{Support Vector Machines: Theory and Applications}
vol. \bseriesno{177}.
\bpublisher{Springer}, \blocation{???}
(\byear{2005})
\end{bbook}
\endbibitem

%%% 14
\bibitem[\protect\citeauthoryear{Hastie et~al.}{2004}]{hastie2004entire}
\begin{barticle}
\bauthor{\bsnm{Hastie}, \binits{T.}},
\bauthor{\bsnm{Rosset}, \binits{S.}},
\bauthor{\bsnm{Tibshirani}, \binits{R.}},
\bauthor{\bsnm{Zhu}, \binits{J.}}:
\batitle{The entire regularization path for the support vector machine}.
\bjtitle{Journal of Machine Learning Research}
\bvolume{5}(\bissue{Oct}),
\bfpage{1391}--\blpage{1415}
(\byear{2004})
\end{barticle}
\endbibitem

%%% 15
\bibitem[\protect\citeauthoryear{Cherkassky and
  Mulier}{2007}]{cherkassky2007learning}
\begin{bbook}
\bauthor{\bsnm{Cherkassky}, \binits{V.}},
\bauthor{\bsnm{Mulier}, \binits{F.M.}}:
\bbtitle{Learning from Data: Concepts, Theory, and Methods}.
\bpublisher{John Wiley \& Sons}, \blocation{???}
(\byear{2007})
\end{bbook}
\endbibitem

%%% 16
\bibitem[\protect\citeauthoryear{Bergstra and
  Bengio}{2012}]{bergstra2012random}
\begin{botherref}
\oauthor{\bsnm{Bergstra}, \binits{J.}},
\oauthor{\bsnm{Bengio}, \binits{Y.}}:
Random search for hyper-parameter optimization.
Journal of machine learning research
\textbf{13}(2)
(2012)
\end{botherref}
\endbibitem

%%% 17
\bibitem[\protect\citeauthoryear{Hsu et~al.}{2003}]{hsu2003practical}
\begin{botherref}
\oauthor{\bsnm{Hsu}, \binits{C.-W.}},
\oauthor{\bsnm{Chang}, \binits{C.-C.}},
\oauthor{\bsnm{Lin}, \binits{C.-J.}}, et al.:
A practical guide to support vector classification.
Taipei
(2003)
\end{botherref}
\endbibitem

%%% 18
\bibitem[\protect\citeauthoryear{Noble}{2006}]{noble2006support}
\begin{barticle}
\bauthor{\bsnm{Noble}, \binits{W.S.}}:
\batitle{What is a support vector machine?}
\bjtitle{Nature biotechnology}
\bvolume{24}(\bissue{12}),
\bfpage{1565}--\blpage{1567}
(\byear{2006})
\end{barticle}
\endbibitem

%%% 19
\bibitem[\protect\citeauthoryear{Sain}{1996}]{sain1996nature}
\begin{botherref}
\oauthor{\bsnm{Sain}, \binits{S.R.}}:
The nature of statistical learning theory.
Taylor \& Francis
(1996)
\end{botherref}
\endbibitem

%%% 20
\bibitem[\protect\citeauthoryear{Sch{\"o}lkopf
  et~al.}{2002}]{scholkopf2002learning}
\begin{bbook}
\bauthor{\bsnm{Sch{\"o}lkopf}, \binits{B.}},
\bauthor{\bsnm{Smola}, \binits{A.J.}},
\bauthor{\bsnm{Bach}, \binits{F.}}, \betal:
\bbtitle{Learning with Kernels: Support Vector Machines, Regularization,
  Optimization, and Beyond}.
\bpublisher{MIT press}, \blocation{???}
(\byear{2002})
\end{bbook}
\endbibitem

%%% 21
\bibitem[\protect\citeauthoryear{Feynman and {Vernon
  Jr.}}{1963}]{Feynman1963118}
\begin{barticle}
\bauthor{\bsnm{Feynman}, \binits{R.P.}},
\bauthor{\bsnm{{Vernon Jr.}}, \binits{F.L.}}:
\batitle{The theory of a general quantum system interacting with a linear
  dissipative system}.
\bjtitle{Annals of Physics}
\bvolume{24},
\bfpage{118}--\blpage{173}
(\byear{1963})
\doiurl{10.1016/0003-4916(63)90068-X}
\end{barticle}
\endbibitem

%%% 22
\bibitem[\protect\citeauthoryear{Allen}{1974}]{allen1974relationship}
\begin{barticle}
\bauthor{\bsnm{Allen}, \binits{D.M.}}:
\batitle{The relationship between variable selection and data agumentation and
  a method for prediction}.
\bjtitle{technometrics}
\bvolume{16}(\bissue{1}),
\bfpage{125}--\blpage{127}
(\byear{1974})
\end{barticle}
\endbibitem

%%% 23
\bibitem[\protect\citeauthoryear{Stone}{1974}]{stone1974cross}
\begin{barticle}
\bauthor{\bsnm{Stone}, \binits{M.}}:
\batitle{Cross-validatory choice and assessment of statistical predictions}.
\bjtitle{Journal of the Royal Statistical Society: Series B (Methodological)}
\bvolume{36}(\bissue{2}),
\bfpage{111}--\blpage{133}
(\byear{1974})
\end{barticle}
\endbibitem

%%% 24
\bibitem[\protect\citeauthoryear{Duarte and Wainer}{2017}]{duarte2017empirical}
\begin{barticle}
\bauthor{\bsnm{Duarte}, \binits{E.}},
\bauthor{\bsnm{Wainer}, \binits{J.}}:
\batitle{Empirical comparison of cross-validation and internal metrics for
  tuning svm hyperparameters}.
\bjtitle{Pattern Recognition Letters}
\bvolume{88},
\bfpage{6}--\blpage{11}
(\byear{2017})
\end{barticle}
\endbibitem

%%% 25
\bibitem[\protect\citeauthoryear{Chang and Lin}{2011}]{chang2011libsvm}
\begin{barticle}
\bauthor{\bsnm{Chang}, \binits{C.-C.}},
\bauthor{\bsnm{Lin}, \binits{C.-J.}}:
\batitle{Libsvm: a library for support vector machines}.
\bjtitle{ACM transactions on intelligent systems and technology (TIST)}
\bvolume{2}(\bissue{3}),
\bfpage{1}--\blpage{27}
(\byear{2011})
\end{barticle}
\endbibitem

%%% 26
\bibitem[\protect\citeauthoryear{Zhu et~al.}{2003}]{zhu20031}
\begin{botherref}
\oauthor{\bsnm{Zhu}, \binits{J.}},
\oauthor{\bsnm{Rosset}, \binits{S.}},
\oauthor{\bsnm{Tibshirani}, \binits{R.}},
\oauthor{\bsnm{Hastie}, \binits{T.}}:
1-norm support vector machines.
Advances in neural information processing systems
\textbf{16}
(2003)
\end{botherref}
\endbibitem

%%% 27
\bibitem[\protect\citeauthoryear{Rahimi and Recht}{2007}]{rahimi2007random}
\begin{botherref}
\oauthor{\bsnm{Rahimi}, \binits{A.}},
\oauthor{\bsnm{Recht}, \binits{B.}}:
Random features for large-scale kernel machines.
Advances in neural information processing systems
\textbf{20}
(2007)
\end{botherref}
\endbibitem

%%% 28
\bibitem[\protect\citeauthoryear{Pham and Pagh}{2013}]{pham2013fast}
\begin{bchapter}
\bauthor{\bsnm{Pham}, \binits{N.}},
\bauthor{\bsnm{Pagh}, \binits{R.}}:
\bctitle{Fast and scalable polynomial kernels via explicit feature maps}.
In: \bbtitle{Proceedings of the 19th ACM SIGKDD International Conference on
  Knowledge Discovery and Data Mining},
pp. \bfpage{239}--\blpage{247}
(\byear{2013})
\end{bchapter}
\endbibitem

%%% 29
\bibitem[\protect\citeauthoryear{Alcal{\'a}-Fdez et~al.}{2009}]{alcala2009keel}
\begin{barticle}
\bauthor{\bsnm{Alcal{\'a}-Fdez}, \binits{J.}},
\bauthor{\bsnm{Sanchez}, \binits{L.}},
\bauthor{\bsnm{Garcia}, \binits{S.}},
\bauthor{\bsnm{Jesus}, \binits{M.J.}},
\bauthor{\bsnm{Ventura}, \binits{S.}},
\bauthor{\bsnm{Garrell}, \binits{J.M.}},
\bauthor{\bsnm{Otero}, \binits{J.}},
\bauthor{\bsnm{Romero}, \binits{C.}},
\bauthor{\bsnm{Bacardit}, \binits{J.}},
\bauthor{\bsnm{Rivas}, \binits{V.M.}}, \betal:
\batitle{Keel: a software tool to assess evolutionary algorithms for data
  mining problems}.
\bjtitle{Soft Computing}
\bvolume{13}(\bissue{3}),
\bfpage{307}--\blpage{318}
(\byear{2009})
\end{barticle}
\endbibitem

%%% 30
\bibitem[\protect\citeauthoryear{Dua and Graff}{2017}]{Dua:2019}
\begin{botherref}
\oauthor{\bsnm{Dua}, \binits{D.}},
\oauthor{\bsnm{Graff}, \binits{C.}}:
{UCI} Machine Learning Repository
(2017).
\url{http://archive.ics.uci.edu/ml}
\end{botherref}
\endbibitem

%%% 31
\bibitem[\protect\citeauthoryear{Deng}{2012}]{deng2012mnist}
\begin{barticle}
\bauthor{\bsnm{Deng}, \binits{L.}}:
\batitle{The mnist database of handwritten digit images for machine learning
  research}.
\bjtitle{IEEE Signal Processing Magazine}
\bvolume{29}(\bissue{6}),
\bfpage{141}--\blpage{142}
(\byear{2012})
\end{barticle}
\endbibitem

\end{thebibliography}

\end{document}